\documentclass[runningheads]{llncs}
\usepackage{graphicx}
\usepackage{amsmath,amssymb} % define this before the line numbering.
\usepackage{color}
\usepackage[width=122mm,left=12mm,paperwidth=146mm,height=193mm,top=12mm,paperheight=217mm]{geometry}
\usepackage{caption}
\DeclareCaptionFont{tiny}{\tiny}

\begin{document}
% \linenumbers
\pagestyle{headings}
\mainmatter
\def\ECCV16SubNumber{866}  % Insert your submission number here

\title{Deep3D: Fully Automatic 2D-to-3D Video Conversion with Deep Convolutional Neural Networks} % Replace with your title

\titlerunning{Deep3D: Automatic 2D-to-3D Video Conversion with DNNs}

\authorrunning{Xie, Girshick, Farhadi}

%\author{Anonymous ECCV submission}
%\institute{Paper ID \ECCV16SubNumber}
\author{Junyuan Xie, Ross Girshick, Ali Farhadi\\jxie@cs.washington.edu, ross.girshick@gmail.com, ali@cs.washington.edu}
\institute{University of Washington}

\maketitle

\begin{figure}[h]
\centering
\includegraphics[width=\textwidth]{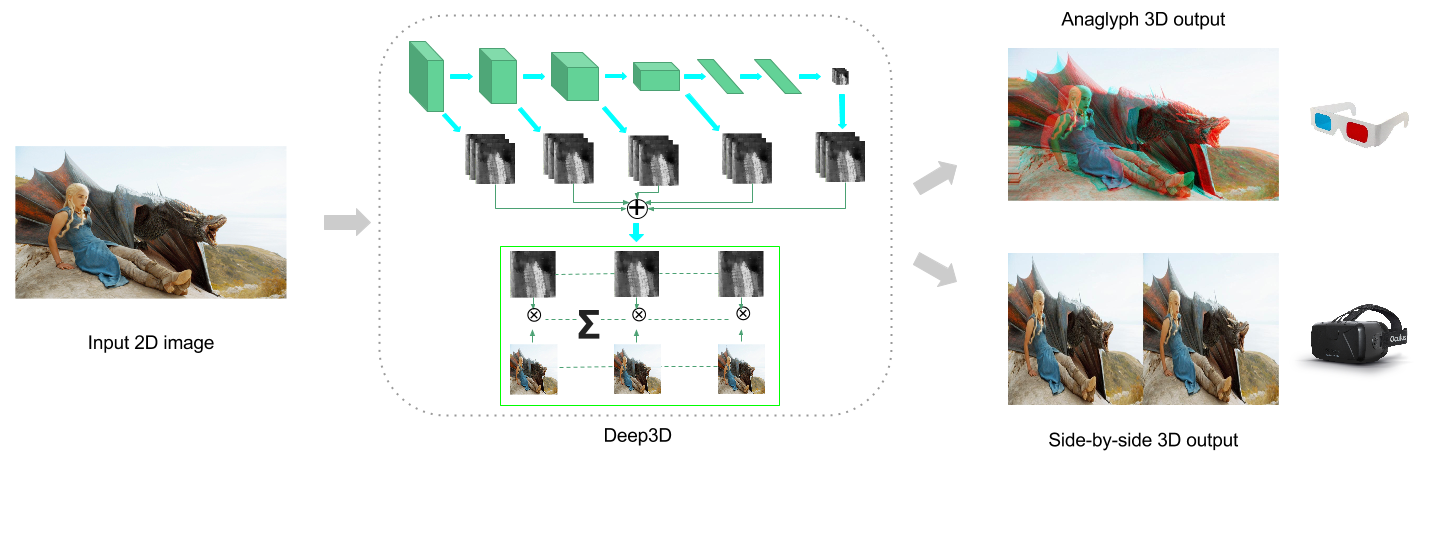}
\vspace{-0.5in}
\caption{We propose Deep3D, a fully automatic 2D-to-3D conversion algorithm that takes 2D images or video frames as input and outputs 3D stereo image pairs.
The stereo images can be viewed with 3D glasses or head-mounted VR displays.
Deep3D is trained directly on stereo pairs from a dataset of 3D movies to minimize the pixel-wise reconstruction error of the right view when given the left view.
Internally, the Deep3D network estimates a probabilistic disparity map that is used by a differentiable depth image-based rendering layer to produce the right view.
Thus Deep3D does not require collecting depth sensor data for supervision.
}
\label{fig:teaser}
\end{figure}

\begin{abstract}
As 3D movie viewing becomes mainstream and Virtual Reality (VR) market emerges, the demand for 3D contents is growing rapidly.
Producing 3D videos, however, remains challenging.
In this paper we propose to use deep neural networks for automatically converting 2D videos and images to stereoscopic 3D format.
In contrast to previous automatic 2D-to-3D conversion algorithms, which have separate stages and need ground truth depth map as supervision, our approach is trained end-to-end directly on stereo pairs extracted from 3D movies.
This novel training scheme makes it possible to exploit orders of magnitude more data and significantly increases performance.
Indeed, Deep3D outperforms baselines in both quantitative and human subject evaluations.
\end{abstract}

\section{Introduction}

3D movies are popular and comprise a large segment of the movie theater market, ranging between 14\% and 21\% of all box office sales between 2010 and 2014 in the U.S. and Canada \cite{mpaa}.
Moreover, the emerging market of Virtual Reality (VR) head-mounted displays will likely drive an increased demand for 3D content.

3D videos and images are usually stored in stereoscopic format.
For each frame, the format includes two projections of the same scene, one of which is exposed to the viewer's left eye and the other to the viewer's right eye, thus giving the viewer the experience of seeing the scene in three dimensions.

%Fully automated generation of 3D videos can significantly lower the cost of supplying this market.

There are two approaches to making 3D movies: shooting natively in 3D or converting to 3D after shooting in 2D.
Shooting in 3D requires costly special-purpose stereo camera rigs.
Aside from equipment costs, there are cinemagraphic issues that may preclude the use of stereo camera rigs.
For example, some inexpensive optical special effects, such as forced perspective\footnote{Forced perspective is an optical illusion technique that makes objects appear larger or smaller than they really are. It breaks down when viewed from another angle, which prevents stereo filming.}, are not compatible with multi-view capture devices.
2D-to-3D conversion offers an alternative to filming in 3D.
Professional conversion processes typically rely on ``depth artists'' who manually create a depth map for each 3D frame.
Standard Depth Image-Based Rendering (DIBR) algorithms can then be used to combine the original frame with the depth map in order to arrive at a stereo image pair \cite{fehn2004depth}.
However, this process is still expensive as it requires intensive human effort.

Each year about 20 new 3D movies are produced.
High production cost is the main hurdle in the way of scaling up the 3D movie industry.
Automated 2D-to-3D conversion would eliminate this obstacle.

In this paper, we propose a fully automated, data-driven approach to the problem of 2D-to-3D video conversion.
%2D-to-3D video conversion is the process by which a video that was originally shot in 2D is converted to a stereoscopic format.
%For each video frame, the stereoscopic format includes two projections of the same scene, one of which is exposed to the viewer's left eye and the other to the viewer's right eye, thus giving the viewer the experience of seeing the scene in three dimensions.
Solving this problem entails reasoning about depth from a single image and synthesizing a novel view for the other eye.
Inferring depth (or disparity) from a single image, however, is a highly under-constrained problem.
In addition to depth ambiguities, some pixels in the novel view correspond to geometry that's not visible in the available view, which causes missing data that must be hallucinated with an in-painting algorithm.

In spite of these difficulties, our intuition is that given the vast number of stereo-frame pairs that exist in already-produced 3D movies it should be possible to train a machine learning model to predict the novel view from the given view.
To that end, we design a deep neural network that takes as input the left eye's view, internally estimates a soft (probabilistic) disparity map, and then renders a novel image for the right eye.
We train our model end-to-end on ground-truth stereo-frame pairs with the objective of directly predicting one view from the other.
The internal disparity-like map produced by the network is computed only in service of creating a good right eye view.
We show that this approach is easier to train for than the alternative of using heuristics to derive a disparity map, training the model to predict disparity directly, and then using the predicted disparity to render the new image.
Our model also performs in-painting implicitly without the need for post-processing.

Evaluating the quality of the 3D scene generated from the left view is non-trivial.
For quantitative evaluations, we use a dataset of 3D movies and report pixel-wise metrics comparing the reconstructed right view and the ground-truth right view.
We also conduct human subject experiments to show the effectiveness of our solution.
We compare our method with the ground-truth and baselines that use state-of-the-art single view depth estimation techniques.
Our quantitative and qualitative analyses demonstrate the benefits of our solution.

\section{Related Work}
Most existing automatic 2D-to-3D conversion pipelines can be roughly divided into two stages.
First, a depth map is estimated from an image of the input view, then a DIBR algorithm combines the depth map with the input view to generate the missing view of a stereo pair.
Early attempts to estimate depth from a single image utilize various hand engineered features and cues including defocus, scattering, and texture gradients \cite{zhuo2009recovery,cozman1997depth}.
% These methods rely on strong assumptions about the scene and photography pipeline rather than using training data and supervised machine learning.
% {\color{red} Eric -- please give an example or two of the strong assumptions.}
% As a result, they perform best in restricted situations in which the assumptions hold.
These methods only rely on one cue.
As a result, they perform best in restricted situations where the particular cue is present.
In contrast, humans perceive depth by seamlessly combining information from multiple sources.

More recent research has moved to learning-based methods \cite{zhang20113d,konrad2013learning,appia2014fully,saxena2009make3d,baig2014im2depth}.
These approaches take single-view 2D images and their depth maps as supervision and try to learn a mapping from 2D image to depth map.
Learning-based methods combine multiple cues and have better generalization, such as recent works that use deep convolutional neural networks (DCNNs) to advance the state-of-the-art for this problem \cite{eigen2014depth,liu2015deep}.
However, collecting high quality image-depth pairs is difficult, expensive, and subject to sensor-dependent constraints.
As a result, existing depth data set mainly consists of a small number of static indoor and, less commonly, outdoor scenes \cite{Silberman:ECCV12,Geiger2013IJRR}.
The lack of volume and variations in these datasets limits the generality of learning-based methods.
Moreover, the depth maps produced by these methods are only an intermediate representation and a separate DIBR step is still needed to generate the final result.

Monocular depth prediction is challenging and we conjecture that performing that task accurately is unnecessary.
Motivated by the recent trend towards training end-to-end differentiable systems \cite{zheng2015conditional,levine2015end}, we propose a method that requires stereo pairs for training and learns to directly predict the right view from the left view.
In our approach, DIBR is implemented using
% the differentiable formulation of Flynn et al. \cite{flynn2015deepstereo}, 
an internal probabilistic disparity representation,
and while it learns something akin to a disparity map the system is allowed to use that internal representation as it likes in service of predicting the novel view.
This flexibility allows the algorithm to naturally handle in-painting.
Unlike 2D image / depth map pairs, there is a vast amount of training data available to our approach since roughly 10 to 20 3D movies have been produced each year since 2008 and each has hundreds of thousands frames.

Our model is inspired by Flynn et al.'s DeepStereo approach \cite{flynn2015deepstereo}, in which they propose to use a probabilistic selection layer to model the rendering process in a differentiable way so that it can be trained together with a DCNN.
Specifically we use the same probabilistic selection layer, but improve upon their approach in two significant ways.
First, their formulation requires two or more calibrated views in order to synthesize a novel view---a restriction that makes it impossible to train from existing 3D movies.
We remove this limitation by restructuring the network input and layout.
Second, their method works on small patches ($28 \times 28$ pixels) which limits the network's receptive field to local structures.
Our approach processes the entire image, allowing large receptive fields that are necessary to take advantage of high-level abstractions and regularities, such as the fact that large people tend to appear close to the camera while small people tend to be far away.

\section{Method}
Previous work on 2D-to-3D conversion usually consists of two steps: estimating an accurate depth map from the left view and rendering the right view with a Depth Image-Based Rendering (DIBR) algorithm.
Instead, we propose to directly regress on the right view with a pixel-wise loss.
Naively following this approach, however, leads to poor results because it does not capture the structure of the task (see Section \ref{sec:ablation}).
Inspired by previous work, we utilize a DIBR process to capture the fact that most output pixels are shifted copies of input pixels.
However, unlike previous work we don't constrain the system to produce an accurate depth map, nor do we require depth maps as supervision for training.
Instead, we propose a model that predicts a probabilistic disparity-like map as an intermediate output and combines it with the input view using a differentiable selection layer that models the DIBR process.
%the differentiable DIBR process introduced in \cite{flynn2015deepstereo}.
During training, the disparity-like map produced by the model is never directly compared to a true disparity map and it ends up serving the dual purposes of representing horizontal disparity and performing in-painting.
Our model can be trained end-to-end thanks to the differentiable selection layer.

%We choose to use L1 loss because it produces sharper images and usually performs better on pixel-wise prediction tasks compared to mean squared error \cite{mathieu2015deep}.
%Specifically, it is beneficial to model the DIBR process explicitly because it captures the fact that most output pixels are shifted copies of input pixels.

\subsection{Model Architecture}
\begin{figure}[!t]
\centering
\includegraphics[width=\textwidth]{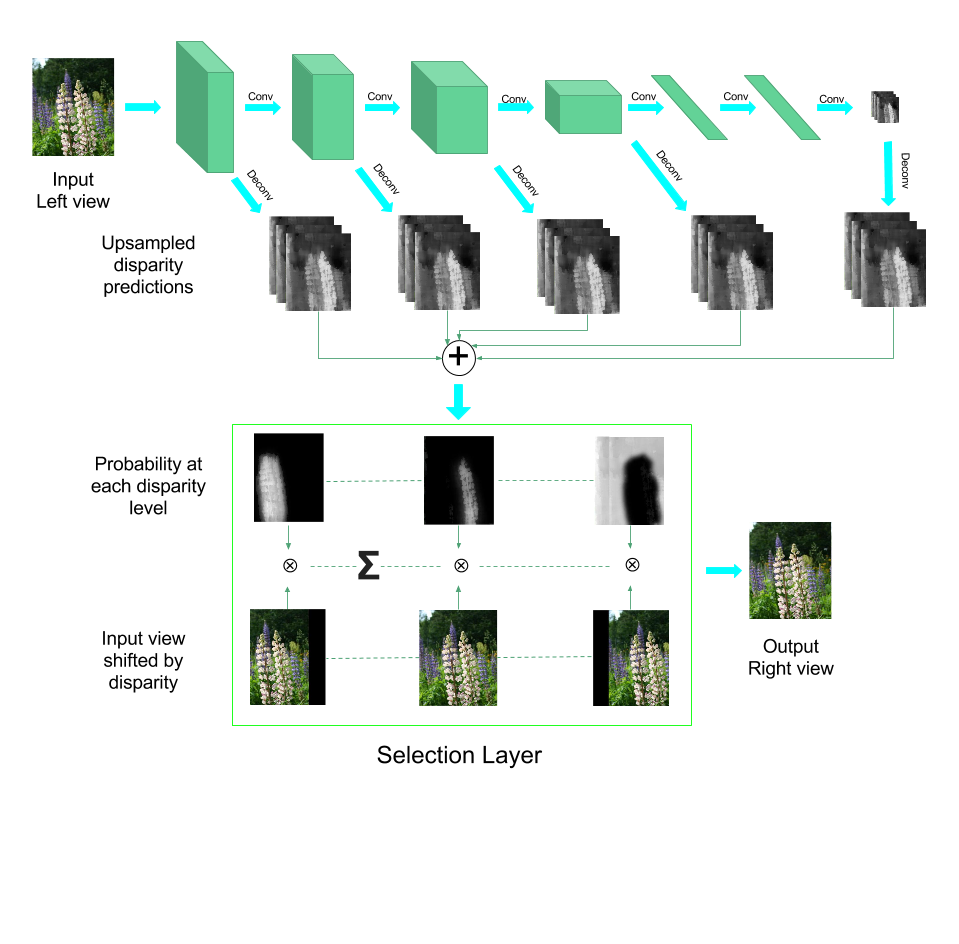}
\vspace{-1in}
\caption{Deep3D model architecture.
Our model combines information from multiple levels and is trained end-to-end to directly generate the right view from the left view.
The base network predicts a probabilistic disparity assignment which is then used by the selection layer to model Depth Image-Based Rendering (DIBR) in a differentiable way.
This also allows implicit in-painting.}
\label{fig:model}
\end{figure}

Recent research has shown that incorporating lower level features benefits pixel wise prediction tasks including semantic segmentation, depth estimation, and optical flow estimation \cite{eigen2014depth,fischer2015flownet}.
Given the similarity between our task and depth estimation, it is natrual to incorporate this idea.
Our network, as shown in Fig. \ref{fig:model}, has a branch after each pooling layer that upsamples the incoming feature maps using so-called ``deconvolution'' layers (i.e., a learned upsampling filter).
The upsampled feature maps from each level are summed together to give a feature representation that has the same size as the input image.
We perform one more convolution on the summed feature representation and apply a softmax transform across channels at each spatial location.
The output of this softmax layer is interpreted as a probabilistic disparity map.
We then feed this disparity map and the left view to the selection layer, which outputs the right view.

\subsubsection{Bilinear Interpolation by Deconvolution}
Similar to \cite{fischer2015flownet} we use ``deconvolutional'' layers to upsample lower layer features maps before feeding them to the final representation.
Deconvolutional layers are implemented by reversing the forward and backward computations of a convolution layer.
Although technically it is still performing convolution, we call it deconvolutional layer following the convention.

We found that initializing the deconvolutional layers to be equivalent to bilinear interpolation can facilitate training.
Specifically, for upsampling by factor $S$, we use a deconvolutional layer with $2S$ by $2S$ kernel, $S$ by $S$ stride, and $S/2$ by $S/2$ padding. The kernel weight $w$ is then initialized with:
\begin{eqnarray}
C &=& \frac{2S - 1 - (S \bmod 2)}{2S}\\
w_{ij} &=& (1 - \vert \frac{i}{S-C}\vert)(1 - \vert \frac{j}{S-C} \vert)
\end{eqnarray}

\subsection{Reconstruction with Selection Layer}
\label{sec:dibr}
\begin{figure}[!t]
\centering
\includegraphics[width=\textwidth]{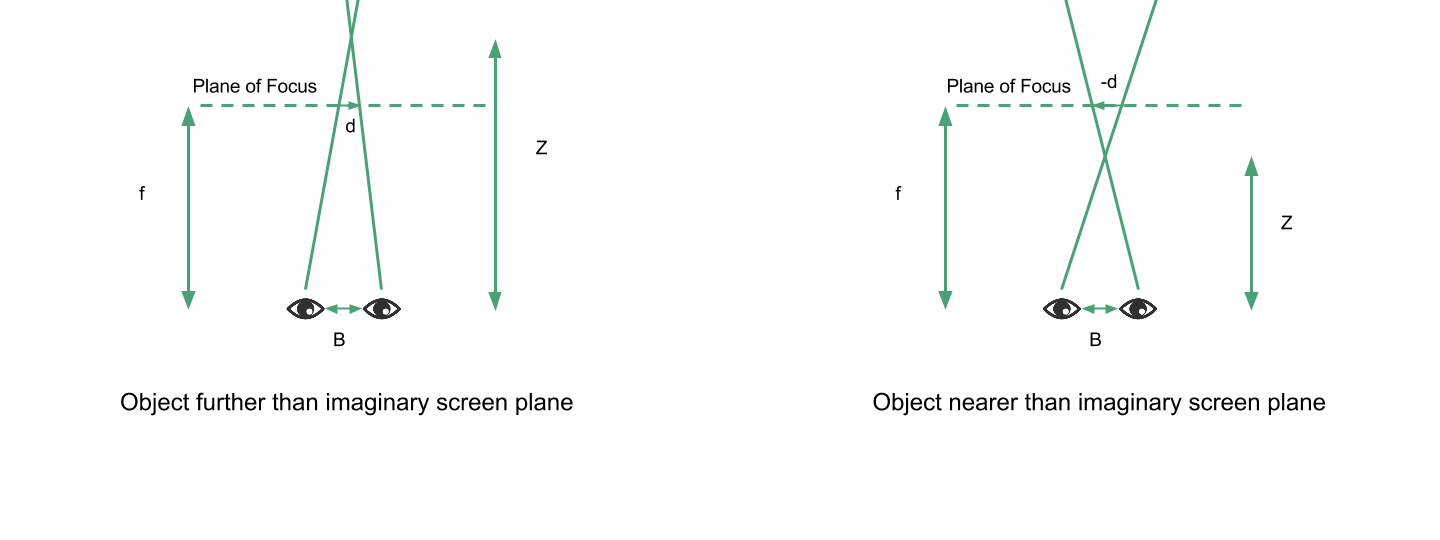}
\vspace{-0.6in}
\caption{Depth to disparity conversion. Given the distance between the eyes $B$ and the distance between the eyes and the plane of focus $f$, we can compute disparity from depth with Eqn. \ref{eqn:disp}. Disparity is negative if object is closer than the plane of focus and positive if it is further away.}
\label{fig:dibr}
\end{figure}
The selection layer models the DIBR step in traditional 2D-to-3D conversion.
In traditional 2D-to-3D conversion, given the left view $I$ and a depth map $Z$, a disparity map $D$ is first computed with
\begin{eqnarray}
D = \frac{B(Z-f)}{Z}
\label{eqn:disp}
\end{eqnarray}
where the baseline $B$ is the distance between the two cameras, $Z$ is the input depth and $f$ is the distance from cameras to the plane of focus. See Fig. \ref{fig:dibr} for illustration.
The right view $O$ is then generated with:
\begin{eqnarray}
O_{i, j+D_{ij}} = I_{i,j}.
\end{eqnarray}
However this is not differentiable with respect to $D$ so we cannot train it together with a deep neural network.
Instead, our network predicts a probability distribution across possible disparity values $d$ at each pixel location $D_{i,j}^d$, where $\sum_d D_{i,j}^d = 1$ for all $i, j$.
We define a shifted stack of the left view as $I_{i,j}^d = I_{i,j-d}$, then the selection layer reconstructs the right view with:
\begin{eqnarray}
O_{i,j} = \sum_d I_{i,j}^d D_{i,j}^d
\end{eqnarray}
This is now differentiable with respect to $D_{i,j}^d$ so we can compute an L1 loss between the output and ground-truth right view $Y$ as the training objective:
\begin{eqnarray}
L = \vert O - Y \vert
\end{eqnarray}
We use L1 loss because recent research has shown that it outperforms L2 loss for pixel-wise prediction tasks \cite{mathieu2015deep}.

\subsection{Scaling Up to Full Resolution}
\label{sec:scale}
Modern movies are usually distributed in at least 1080p resolution, which has 1920 pixel by 1080 pixel frames.
In our experiments, We reduce input frames to 432 by 180 to preserve aspect ratio and save computation time.
As a result, the generated right view frames will only have a resolution of 432 by 180, which is unacceptably low for movie viewing.

To address this issue, we first observe that the disparity map usually has much less high-frequency content than the original color image.
Therefore we can scale up the predicted disparity map and couple it with the original high resolution left view to render a full resolution right view.
The right view rendered this way has better image quality compared to the naively 4x-upsampled prediction.

\section{Dataset}
Since Deep3D can be trained directly on stereo pairs without ground-truth depth maps as supervision,
we can take advantage of the large volume of existing stereo videos instead of using traditional scene depth datasets like KITTI \cite{Geiger2013IJRR} and NYU Depth \cite{Silberman:ECCV12}.
We collected 27 non-animation 3D movies produced in recent years and randomly partitioned them to 18 for training and 9 for testing.
Our dataset contains around 5 million frames while KITTI and NYU Depth only provide several hundred frames.
During training, each input left frame is resized to 432 by 180 pixels and a crop of size 384 by 160 pixels is randomly selected from the frame.
The target right frame undergoes the same transformations.
We do not use horizontal flipping.

\section{Experiments}
In our main experiments we use a single frame at a time as input without exploiting temporal information.
This choice ensures fair comparison to single-frame baseline algorithms and also allows applying trained models to static photos in addition to videos.
However, it is natural to hypothesize that motion provides important cues for depth, thus we also conducted additional experiments that use consecutive RGB frames and computed optical flow as input, following \cite{wang2015towards}.
% Somewhat surprisingly, our model achieves similar quantitative results both with and without motion.
These results are discussed in Section \ref{sec:tempo}.

\subsection{Implementation Details}
For quantitative evaluation we use the non-upsampled output size of 384 by 160 pixels.
For qualitative and human subject evaluation we upsample the output by a factor of 4 using the method described in Section \ref{sec:scale}.
Our network is based on VGG16, which is a large convolutional network trained on ImageNet \cite{simonyan2014very}.
We initialize the main branch convolutional layers (colored green in Fig.\ref{fig:model}) with VGG16 weight and initialize all other weights with normal distribution with a standard deviation of 0.01.

To integrate information from lower level features, we create a side branch after each pooling layer by applying batch normalization \cite{ioffe2015batch} followed by a $3 \times 3$ convolution layer. This is then followed by a deconvolution layer initialized to be equivalent to bilinear upsampling.
The output dimensions of the deconvolution layers match the final prediction dimensions.
We use batch normalization to connect pretrained VGG16 layers to randomly initialized layers because it solves the numerical instability problem caused by VGG16's large and non-uniform activation magnitude.

We also connect the top VGG16 convolution layer feature to two randomly initialized fully connected layers (colored blue in Fig.\ref{fig:model}) with 4096 hidden units followed by a linear layer.
We then reshape the output of the linear layer to 33 channels of 12 by 5 feature maps which is then fed to a deconvolution layer.
We then sum across all up sampled feature maps and do a convolution to get the final feature representation.
The representation is then fed to the selection layer.
The selection layer interprets this representation as the probability over empty or disparity -15 to 16 (a total of 33 channels).

In all experiments Deep3D is trained with a mini-batch size of 64 for $100,000$ iterations in total.
The initial learning rate is set to 0.002 and reduce it by a factor of 10 after every $20,000$ iterations.
No weight decay is used
and dropout with rate 0.5 is only applied after the fully connected layers.
Training takes two days on one NVidia GTX Titan X GPU.
Once trained, Deep3D can reconstruct novel right views at more than 100 frames per second.
Our implementation is based on MXNet \cite{chen2015mxnet} and available for download at \url{https://github.com/piiswrong/deep3d}.

\subsection{Comparison Algorithms}
We used three baseline algorithms for comparison:
\begin{enumerate}
\item Global Disparity: the right view is computed by shifting the left view with a global disparity $\delta$ that is determined by minimizing Mean Absolution Error (MAE) on the validation set.
\item The DNN-based single image depth estimation algorithm of Eigen et al. \cite{eigen2014depth} plus a standard DIBR method as described in Section \ref{sec:dibr}.
\item Ground-truth stereo pairs. We only show the ground-truth in human subject studies since in quantitative evaluations it always gives zero error.
\end{enumerate}

To the best of our knowledge, Deep3D is the first algorithm that can be trained directly on stereo pairs,
while all previous methods requires ground-truth depth map for training.
For this reason, we cannot retrain comparison algorithms on our 3D movie data set.
Instead, we take the model released by Eigen et al. \cite{eigen2014depth} and evaluate it on our test set.
While it is a stretch to hope that a model trained on NYU Depth will generalize well to 3D movies, this fact underscores a principal strength of our approach: by directly training on stereo pairs, we can exploit vastly more training data.

Because \cite{eigen2014depth} predicts depth rather then disparity, we need to convert depth to disparity with Eqn. \ref{eqn:disp} for rendering with DIBR.
However, \cite{eigen2014depth} does not predict the distance to the plane of focus $f$, a quantity that is unknown and varies across shots due to zooming.
The interpupillary distance $B$ is also unknown, but it is fixed across shots.
The value of $B$ and $f$ can be determined in two ways:
\begin{enumerate}
\item Optimize for MAE on the validation set and use fixed values for $B$ and $f$ across the whole test set.
This approach corresponds to the lower bound of \cite{eigen2014depth}'s performance.
\item Fix $B$ across the test set, but pick the $f$ that gives the lowest MAE for each \emph{test} frame.
This corresponds to having access to oracle plane of focus distance and thus the upper bound on \cite{eigen2014depth}'s performance.
\end{enumerate}

We do both and report them as two separate baselines, \cite{eigen2014depth} and \cite{eigen2014depth} + Oracle.
For fair comparisons, we also do this optimization for Deep3D's predictions and report the performance of Deep3D and Deep3D + Oracle.

\subsection{Results}
\begin{table}[!ht]
\begin{center}
\caption{Deep3D evaluation. We compare pixel-wise reconstruction error for each method using Mean Absolute Error (MAE) as metric.}
\label{table:objeval}
\begin{tabular}{lc}
\hline\noalign{\smallskip}
Method & MAE\\
\noalign{\smallskip}
\hline
\noalign{\smallskip}
Global Disparity                             & 7.75\\
\cite{eigen2014depth}                        & 7.75\\
Deep3D (ours)                                & \textbf{6.87}\\
\hline
\cite{eigen2014depth} + Oracle \hspace{0.8in}               & 6.31\\
Deep3D + Oracle                              & \textbf{5.47}\\
\hline
\end{tabular}
\end{center}
\end{table}

\begin{table}[t]
\begin{center}
\caption{Human Subject Evaluation. Each entry represents the frequency of the row method being preferred to the column method by human subjects.
Note that 66\% of times subjects prefer Deep3D to \cite{eigen2014depth} and 24\% of the times Deep3D is preferred over the ground truth.}
\label{table:subjeval}
\begin{tabular}{l|c|c|c|c}
\hline\noalign{\smallskip}
  & Global Disparity & \cite{eigen2014depth} + Oracle & Deep3D (ours) & Ground Truth\\
\noalign{\smallskip}
\hline
\noalign{\smallskip}
Global Disparity                    &   N/A &   26.94\% &   25.42\% &   7.88\%  \\
\cite{eigen2014depth} + Oracle      &   73.06\% &   N/A &   33.92\% &   10.27\% \\
Deep3D (ours)                       &   74.58\% &   66.08\% &   N/A &   24.48\% \\
Ground Truth                        &   92.12\% &   89.73\% &   75.52\% &   N/A \\
\hline
\end{tabular}
\end{center}
\end{table}
\subsubsection{Quantitative Evaluation}
%During testing, we use movies that are not seen in training set.
For quantitative evaluation, we compute Mean Absolute Error (MAE) as:  
\begin{eqnarray}
MAE = \frac{1}{HW}\vert y - g(x) \vert, \\
\end{eqnarray}
where $x$ is the left view, $y$ is the right view, $g(\cdot)$ is the model, and $H$ and $W$ are height and width of the image respectively.
The results are shown in Table \ref{table:objeval}.
We observe that Deep3D outperforms baselines with and without oracle distance of focus plane.

\begin{figure}[t]
\includegraphics[height=0.23\textheight]{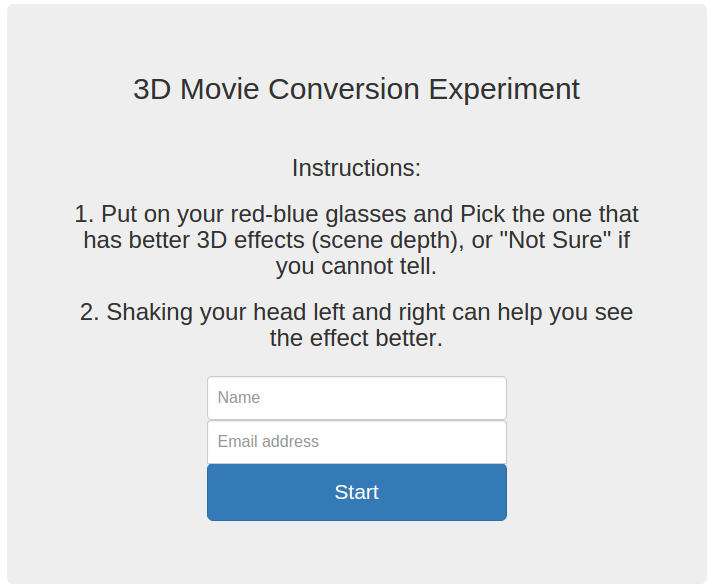}
\quad
\includegraphics[height=0.23\textheight]{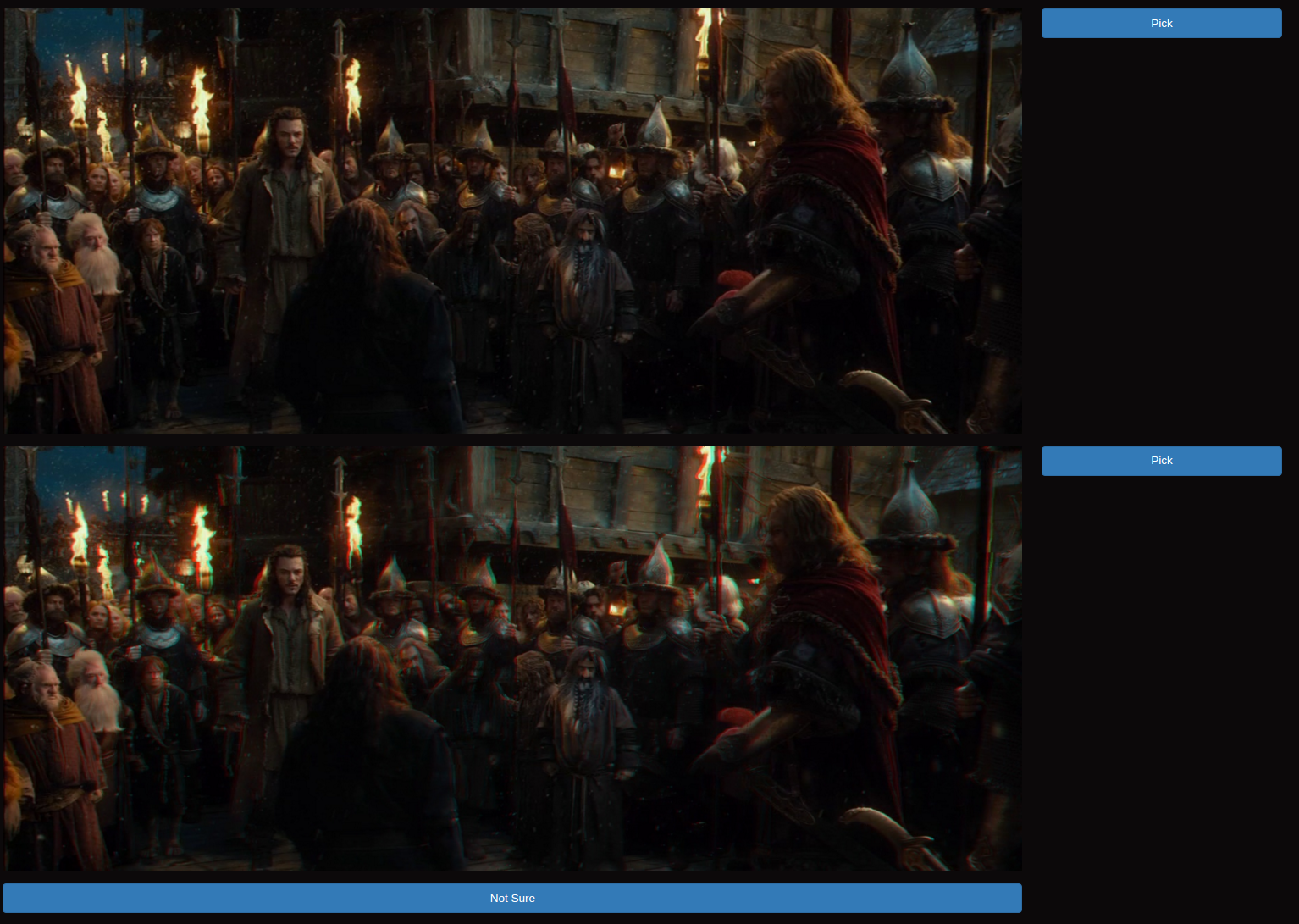}
\caption{Human subject study setup. Each subject is shown 50 pairs of 3D anaglyph images.
Each pair consists of the same scene generated by 2 randomly selected methods.
The subjects are instructed to wear red-blue 3D glasses and pick the one with better 3D effects or ``Not Sure'' if they cannot tell.
The study result is shown in Table \ref{table:subjeval}}
\label{fig:study}
\end{figure}

\subsubsection{Qualitative Evaluation}
\begin{figure}[!p]
\begin{tabular}{c}
\includegraphics[width=0.95\textwidth]{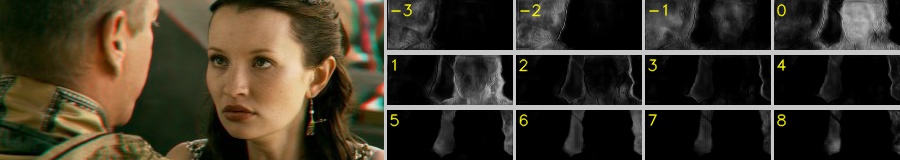}\\
\includegraphics[width=0.95\textwidth]{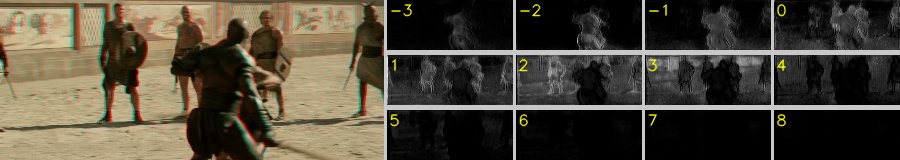}\\
\includegraphics[width=0.95\textwidth]{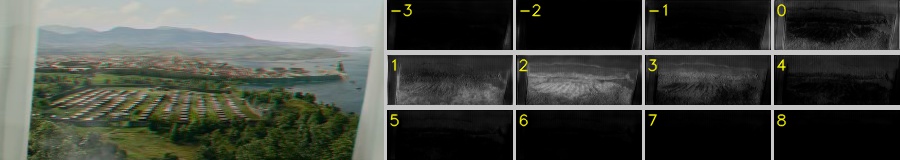}\\
\includegraphics[width=0.95\textwidth]{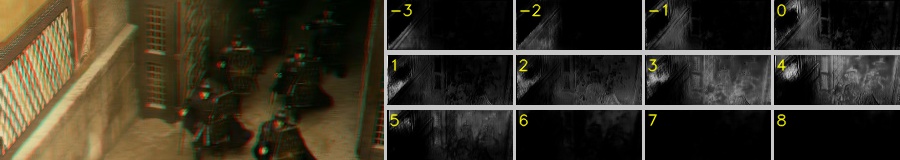}\\
\includegraphics[width=0.95\textwidth]{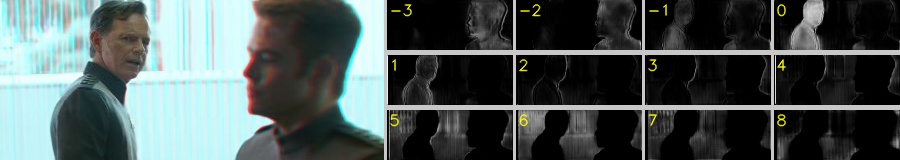}\\
\includegraphics[width=0.95\textwidth]{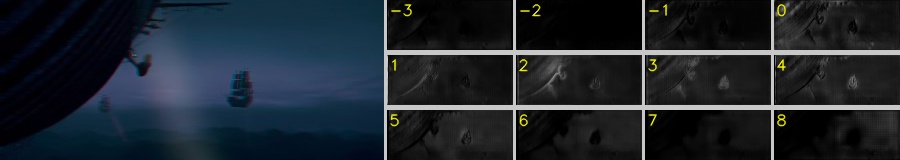}\\
\includegraphics[width=0.95\textwidth]{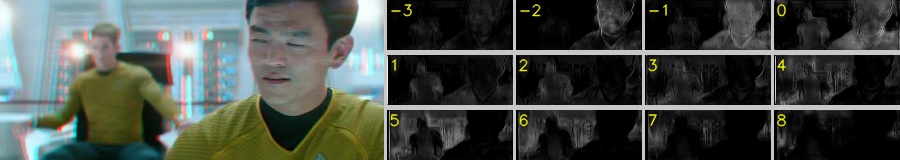}\\
\includegraphics[width=0.95\textwidth]{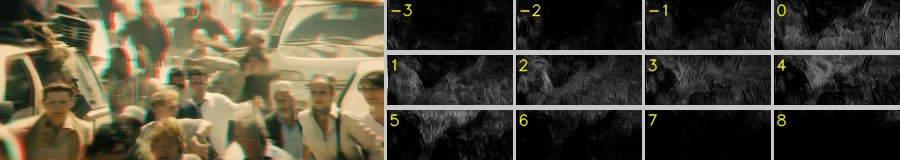}\\
\end{tabular}
\captionsetup{font=tiny}
\vspace{-0.1in}
\caption{Qualitative results.
Column one shows an anaglyph of the predicted 3D image (best viewed in color with red-blue 3D glasses).
Each anaglyph is followed by 12 heat maps of disparity channels -3 to 8 (closer to far).
In the first example, the man is closer and appears in the first 3 channels while the woman is further away and appears in 4th-5th channels; the background appears in the last 4 channels.
In the second example, the person seen from behind is closer than the other 4 people fighting him.
In the third example, the window frame appears in the first 3 channels while the distant outside scene gradually appears in the following channels.}
\label{fig:depth}
\end{figure}

To better understand the proposed method, we show qualitative results in Fig. \ref{fig:depth}.
Each entry starts with a stereo pair predicted by Deep3D shown in anaglyph, followed by 12 channels of internal soft disparity assignment, ordered from near (-3) to far (+8).
We observe that Deep3D is able to infer depth from multiple cues including size, occlusion, and geometric structure.

\begin{figure}[t]
\includegraphics[width=\textwidth]{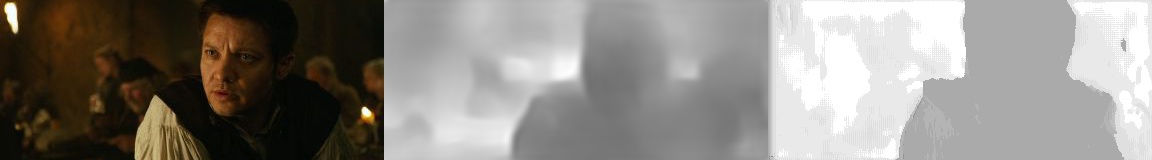}
\includegraphics[width=\textwidth]{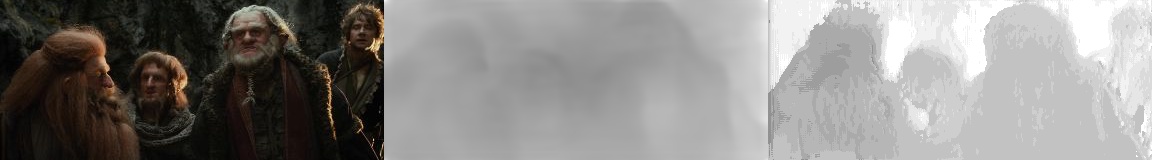}
\includegraphics[width=\textwidth]{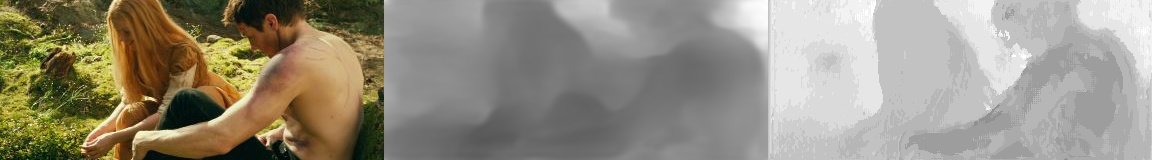}
\caption{Comparison between \cite{eigen2014depth} and Deep3D. The first column shows the input image. The second column shows the prediction of \cite{eigen2014depth} and the third column shows Deep3D's prediction. This figure shows that Deep3D is better at delineating people and figuring out their distance from the camera.}
\label{fig:comp}
\end{figure}

We also compare Deep3D's internal disparity maps (column 3) to \cite{eigen2014depth}'s depth predictions (column 2) in \ref{fig:comp}.
This figure demonstrates that Deep3D is better at delineating people and figuring out their distance from the camera.

Note that the disparity maps generated by Deep3D tend to be noisy at image regions with low horizontal gradient, however this does not affect the quality of the final reconstruction because if a row of pixels have the same value, any disparity assignment would give the same reconstruction.
Disparity prediction only needs to be accurate around vertical edges and we indeed observe that Deep3D tends to focus on such regions.

\subsubsection{Human Subject Evaluation}

We also conducted a human subject study to evaluate the visual quality of the predictions of different algorithms.
We used four algorithms for this experiment: Global Disparity, \cite{eigen2014depth} + Oracle, Deep3D without Oracle, and the ground-truth\footnote{
\cite{eigen2014depth} without Oracle and Deep3D + Oracle are left out due to annotator budget.
Note that a change in average scene depth only pushes a scene further away or pull it closer and usually doesn't affect the perception of depth variation in the scene.}.

For the human subject study, we randomly selected 500 frames from the test set.
Each annotator is shown a sequence of trials.
In each trial, the annotator sees two anaglyph 3D images, which are reconstructed from the same 2D frame by two algorithms, and is instructed to wear red-blue 3D glasses and pick the one with better 3D effects or select ``not sure'' if they are similar.
The interface for this study is shown in Fig. \ref{fig:study}.
Each annotator is given 50 such pairs and we collected decisions on all $C_4^2 500$ pairs from 60 annotators.
% ######## {\color{red} Don't say roughly 60, give the exact number} ###########

Table \ref{table:subjeval} shows that Deep3D outperforms the naive Global Disparity baseline by a $49\%$ margin and outperforms \cite{eigen2014depth} + Oracle by a $32\%$ margin. 
When facing against the ground truth, Deep3D's prediction is preferred $24.48\%$ of the time while \cite{eigen2014depth} + Oracle is only preferred $10.27\%$ of the time and Global Disparity baseline is preferred $7.88\%$ of the time.

\subsection{Algorithm Analysis}

\subsubsection{Ablation Study}
\label{sec:ablation}
\begin{table}[!ht]
\begin{center}
\caption{Ablation studies. We evaluate different components of Deep3D by removing them from the model to further understand the contribution of each component. Note that removing lower level features and selection layer both result in performance drop.}
\label{table:ablation}
\begin{tabular}{lc}
\hline\noalign{\smallskip}
Method & MAE\\
\noalign{\smallskip}
\hline
\noalign{\smallskip}
Deep3D w/o lower level feature              & 8.24\\
Deep3D w/o direct training on stereo pairs \hspace{0.5in}  & 7.29\\
Deep3D w/o selection layer                  & 7.01\\
Deep3D                                      & 6.87\\
\hline
\end{tabular}
\end{center}
\end{table}

To understand the contribution of each component of the proposed algorithm, we show the performance of Deep3D with parts removed in Tab. \ref{table:ablation}.
In Deep3D w/o lower level feature we show the performance of Deep3D without branching off from lower convolution layers.
The resulting network only has one feed-forward path that consists of 5 convolution and pooling module and 2 fully connected layers.
We observe that the performance significantly decreases compared to the full method.

In Deep3D w/o direct training on stereo pairs we show the performance of training on disparity maps generated from stereo pairs by block matching algorithm \cite{hirschmuller2008stereo} instead of directly training on stereo pairs.
The predicted disparity maps are then fed to DIBR method to render the right view.
This approach results in decreased performance and demonstrates the effectiveness of Deep3D's end-to-end training scheme.

We also show the result from directly regressing on the novel view without internal disparity representation and selection layer.
Empirically this also leads to decreased performance, demonstrating the effectiveness of modeling the DIBR process.

\subsubsection{Temporal Information}
\label{sec:tempo}

\begin{table}[!ht]
\begin{center}
\caption{Temporal information. We incorporate temporal information by extending the input to include multiple consecutive RGB frames or optical flow frames. We observe that additional temporal information leads to performance gains.}
\label{table:tempo}
\begin{tabular}{lc}
\hline\noalign{\smallskip}
Method & MAE\\
\hline
Deep3D with 5 RGB frames                                   & 6.81\\
Deep3D with 1 RGB frames and 5 optical flow frames \hspace{0.3in}         & 6.86\\
Deep3D                                                     & 6.87\\
\hline
\end{tabular}
\end{center}
\end{table}

In our main experiment and evaluation we only used one still frame of RGB image as input.
We made this choice for fair comparisons and more general application domains.
Incorporating temporal information into Deep3D can be handled in two ways:
use multiple consecutive RGB frames as input to the network,
or provide temporal information through optical flow frames similar to \cite{wang2015towards}.

We briefly explored both directions and found moderate performance improvements in terms of pixel-wise metrics.
We believe more effort along this direction, such as model structure adjustment, hyper-parameter tuning, and explicit modeling of time will lead to larger performance gains at the cost of restricting application domain to videos only.
% We have three hypothesizes for this phenomenon:
% 1. temporal information is not needed for this task,
% 2. effectively incorporating it is non-trivial and naive approaches does not work,
% 3. pixel wise metrics have lower bounds due to inherent randomness in the data set, human subject evaluation is needed to differentiate between high performance algorithms.
% We think that 1) is highly unlikely and 2 \& 3 are interesting future research directions.

% \input{discussion}
\section{Conclusions}
In this paper we proposed a fully automatic 2D-to-3D conversion algorithm based on deep convolutional neural networks.
Our method is trained end-to-end on stereo image pairs directly, thus able to exploit orders of magnitude more data than traditional learning based 2D-to-3D conversion methods.
Quantitatively, our method outperforms baseline algorithms.
In human subject study stereo images generated by our method are consistently preferred by subjects over results from baseline algorithms.
When facing against the ground truth, our results have a higher chance of confusing %fooling? need a better word
subjects than baseline results.

In our experiment and evaluations we only used still images as input while ignoring temporal information from video.
The benefit of this design is that the trained model can be applied to not only videos but also photos.
However, in the context of video conversion, it is likely that taking advantage of temporal information can improve performance.
We briefly experimented with this idea but found little quantitative performance gain.
We conjecture this may be due to the complexity of effectively incorporating temporal information.
We believe this is an interesting direction for future research.

\bibliographystyle{splncs}
\bibliography{egbib}

\begin{thebibliography}{10}

\bibitem{mpaa}
{Motion Picture Association of America}:
\newblock Theatrical market statistics.
\newblock (2014)

\bibitem{fehn2004depth}
Fehn, C.:
\newblock Depth-image-based rendering (dibr), compression, and transmission for
  a new approach on 3d-tv.
\newblock In: Electronic Imaging 2004, International Society for Optics and
  Photonics (2004)  93--104

\bibitem{zhuo2009recovery}
Zhuo, S., Sim, T.:
\newblock On the recovery of depth from a single defocused image.
\newblock In: Computer Analysis of Images and Patterns, Springer (2009)
  889--897

\bibitem{cozman1997depth}
Cozman, F., Krotkov, E.:
\newblock Depth from scattering.
\newblock In: Computer Vision and Pattern Recognition, 1997. Proceedings., 1997
  IEEE Computer Society Conference on, IEEE (1997)  801--806

\bibitem{zhang20113d}
Zhang, L., V{\'a}zquez, C., Knorr, S.:
\newblock 3d-tv content creation: automatic 2d-to-3d video conversion.
\newblock Broadcasting, IEEE Transactions on \textbf{57}(2) (2011)  372--383

\bibitem{konrad2013learning}
Konrad, J., Wang, M., Ishwar, P., Wu, C., Mukherjee, D.:
\newblock Learning-based, automatic 2d-to-3d image and video conversion.
\newblock Image Processing, IEEE Transactions on \textbf{22}(9) (2013)
  3485--3496

\bibitem{appia2014fully}
Appia, V., Batur, U.:
\newblock Fully automatic 2d to 3d conversion with aid of high-level image
  features.
\newblock In: IS\&T/SPIE Electronic Imaging, International Society for Optics
  and Photonics (2014)  90110W--90110W

\bibitem{saxena2009make3d}
Saxena, A., Sun, M., Ng, A.Y.:
\newblock Make3d: Learning 3d scene structure from a single still image.
\newblock Pattern Analysis and Machine Intelligence, IEEE Transactions on
  \textbf{31}(5) (2009)  824--840

\bibitem{baig2014im2depth}
Baig, M.H., Jagadeesh, V., Piramuthu, R., Bhardwaj, A., Di, W., Sundaresan, N.:
\newblock Im2depth: Scalable exemplar based depth transfer.
\newblock In: Applications of Computer Vision (WACV), 2014 IEEE Winter
  Conference on, IEEE (2014)  145--152

\bibitem{eigen2014depth}
Eigen, D., Puhrsch, C., Fergus, R.:
\newblock Depth map prediction from a single image using a multi-scale deep
  network.
\newblock In: Advances in neural information processing systems. (2014)
  2366--2374

\bibitem{liu2015deep}
Liu, F., Shen, C., Lin, G.:
\newblock Deep convolutional neural fields for depth estimation from a single
  image.
\newblock In: Proceedings of the IEEE Conference on Computer Vision and Pattern
  Recognition. (2015)  5162--5170

\bibitem{Silberman:ECCV12}
Nathan~Silberman, Derek~Hoiem, P.K., Fergus, R.:
\newblock Indoor segmentation and support inference from rgbd images.
\newblock In: ECCV. (2012)

\bibitem{Geiger2013IJRR}
Geiger, A., Lenz, P., Stiller, C., Urtasun, R.:
\newblock Vision meets robotics: The kitti dataset.
\newblock International Journal of Robotics Research (IJRR) (2013)

\bibitem{zheng2015conditional}
Zheng, S., Jayasumana, S., Romera-Paredes, B., Vineet, V., Su, Z., Du, D.,
  Huang, C., Torr, P.H.:
\newblock Conditional random fields as recurrent neural networks.
\newblock In: Proceedings of the IEEE International Conference on Computer
  Vision. (2015)  1529--1537

\bibitem{levine2015end}
Levine, S., Finn, C., Darrell, T., Abbeel, P.:
\newblock End-to-end training of deep visuomotor policies.
\newblock arXiv preprint arXiv:1504.00702 (2015)

\bibitem{flynn2015deepstereo}
Flynn, J., Neulander, I., Philbin, J., Snavely, N.:
\newblock Deepstereo: Learning to predict new views from the world's imagery.
\newblock arXiv preprint arXiv:1506.06825 (2015)

\bibitem{fischer2015flownet}
Fischer, P., Dosovitskiy, A., Ilg, E., H{\"a}usser, P., Haz{\i}rba{\c{s}}, C.,
  Golkov, V., van~der Smagt, P., Cremers, D., Brox, T.:
\newblock Flownet: Learning optical flow with convolutional networks.
\newblock arXiv preprint arXiv:1504.06852 (2015)

\bibitem{mathieu2015deep}
Mathieu, M., Couprie, C., LeCun, Y.:
\newblock Deep multi-scale video prediction beyond mean square error.
\newblock arXiv preprint arXiv:1511.05440 (2015)

\bibitem{wang2015towards}
Wang, L., Xiong, Y., Wang, Z., Qiao, Y.:
\newblock Towards good practices for very deep two-stream convnets.
\newblock arXiv preprint arXiv:1507.02159 (2015)

\bibitem{simonyan2014very}
Simonyan, K., Zisserman, A.:
\newblock Very deep convolutional networks for large-scale image recognition.
\newblock arXiv preprint arXiv:1409.1556 (2014)

\bibitem{ioffe2015batch}
Ioffe, S., Szegedy, C.:
\newblock Batch normalization: Accelerating deep network training by reducing
  internal covariate shift.
\newblock arXiv preprint arXiv:1502.03167 (2015)

\bibitem{chen2015mxnet}
Chen, T., Li, M., Li, Y., Lin, M., Wang, N., Wang, M., Xiao, T., Xu, B., Zhang,
  C., Zhang, Z.:
\newblock Mxnet: A flexible and efficient machine learning library for
  heterogeneous distributed systems.
\newblock arXiv preprint arXiv:1512.01274 (2015)

\bibitem{hirschmuller2008stereo}
Hirschm{\"u}ller, H.:
\newblock Stereo processing by semiglobal matching and mutual information.
\newblock Pattern Analysis and Machine Intelligence, IEEE Transactions on
  \textbf{30}(2) (2008)  328--341

\end{thebibliography}
\end{document}

\title{Supplementary Material for `Deep3D: Fully Automatic 2D-to-3D Video Conversion with Deep Convolutional Neural Networks'}
\date{}
\maketitle

\section{3D Image Demo}
We provide high resolution 3D images generated by Deep3D in three formats:
\begin{enumerate}
\item Anaglyph 3D: this format uses red vs green-blue channels to display the left and right eye views respectively.
It can be viewed with red-blue glasses.
\item Side-by-Side 3D: this format displays the two views side-by-side each with half width. It can be viewed with 3D media players on Oculus or 3D TVs.
\item GIF: this format displays the two views alternatively in a gif image. It can be viewed on normal screens without special glasses.
\end{enumerate}

\section{3D Movie Clip}
We provide 3D movie clips generated by Deep3D in two formats: anaglyph and side-by-side. The videos are encoded with X264 codec and should be playable with the VLC player.

\section{Dataset}
We used 27 3D movies in our experiment; 18 for training and 9 for testing. In this section we provide a list of used movies.
\subsection{Training}
\begin{enumerate}
\item R.I.P.D. (2013)
\item Hercules (2014) 
\item Jupiter Ascending (2015)
\item Maleficent (2014) 
\item Dredd (2012)
\item The Young and Prodigious T.S. Spivet (2013)
\item The Walk (2015) 
\item Guardians of the Galaxy (2014)
\item Exodus: Gods and Kings (2014) 
\item Dracula 3D (2012)
\item Mad Max: Fury Road (2015)
\item X-Men: Days of Future Past (2014)
\item Iron Man 3 (2013) 
\item Captain America: The Winter Soldier (2014)
\item 300: Rise of an Empire (2014)
\item Thor: The Dark World (2013)
\item Demonic (2015) 
\item The Wolverine (2013)
\end{enumerate}

\subsection{Testing}
\begin{enumerate}
\item The Hobbit: The Desolation of Smaug (2013)
\item Gravity (2013)
\item World War Z (2013)
\item Star Trek Into Darkness (2013) 
\item Pompeii (2014) 
\item Pan (2015)
\item Resident Evil: Afterlife (2010)
\item I, Frankenstein (2014) 
\item Hansel \& Gretel: Witch Hunters (2013)
\end{enumerate}